%% file: main.tex
\documentclass[a4paper,fleqn]{cas-dc}

\usepackage[numbers]{natbib}
\usepackage{amsmath,amssymb,amsfonts} 
\usepackage{booktabs}
\usepackage{xspace}
\usepackage{algorithm,algorithmic}
\usepackage{color}
\usepackage{dsfont}
\usepackage{tabularx}
\usepackage{multirow}
\usepackage{bbm}
\usepackage{hyperref}
\usepackage{times}
\usepackage{balance}

\usepackage{subcaption}
\def\tsc#1{\csdef{#1}{\textsc{\lowercase{#1}}\xspace}}
\tsc{WGM}
\tsc{QE}
\input{notation.tex}

\begin{document}
\let\WriteBookmarks\relax
\def\floatpagepagefraction{1}
\def\textpagefraction{.001}

\shorttitle{Improving rule mining via embedding-based link prediction}    

\shortauthors{Kouagou, Yilmaz, Dumontier, Ngonga Ngomo}  

\title [mode = title]{Discovering the unknown: Improving rule mining via embedding-based link prediction}  



%

\author[1]{N'Dah Jean Kouagou}[orcid=0000-0002-4217-897X]

\cormark[1]


\ead{ndah.jean.kouagou@upb.de}



\affiliation[1]{organization={Paderborn University},
            addressline={Warburger Str.}, 
            city={Paderborn},
            postcode={33098}, 
            state={NRW},
            country={Germany}}

\author[2]{Arif Yilmaz}


\ead{a.yilmaz@maastrichtuniversity.nl}



\affiliation[2]{organization={Maastricht University},
            addressline={MD}, 
            city={Maastricht},
            postcode={P.O. Box 616 6200 MD}, 
            state={Limburg},
            country={The Netherlands}}

\author[2]{Michel Dumontier}


\ead{michel.dumontier@maastrichtuniversity.nl}

\author[1]{Axel-Cyrille {Ngonga Ngomo}}[orcid=0000-0001-7112-3516]


\ead{axel.ngonga@upb.de}

\cortext[1]{Corresponding author}



\begin{abstract}
Rule mining on knowledge graphs allows for explainable link prediction. Contrarily, embedding-based methods for link prediction are well known for their generalization capabilities, but their predictions are not interpretable. Several approaches combining the two families have been proposed in recent years. The majority of the resulting hybrid approaches are usually trained within a unified learning framework, which often leads to convergence issues due to the complexity of the learning task. In this work, we propose a new way to combine the two families of approaches. Specifically, we enrich a given knowledge graph by means of its pre-trained entity and relation embeddings before applying rule mining systems on the enriched knowledge graph. To validate our approach, we conduct extensive experiments on seven benchmark datasets. An analysis of the results generated by our approach suggests that we discover new valuable rules on the enriched graphs. We provide an open source implementation of our approach as well as pretrained models and datasets at \url{https://github.com/Jean-KOUAGOU/EnhancedRuleLearning}.
\end{abstract}



\maketitle
\section{Introduction}
Knowledge graphs are gaining increasing attention due to their wide range of applications \cite{hamad2018food,krishnan2018making}, including question answering \cite{hixon2015learning,lukovnikov2017neural}, and drug discovery \cite{bonner2021review,maclean2021knowledge}. They describe facts in the form of triples ($h$, $r$, $t$), where $h$ is the head entity, $r$ the relation, and $t$ the tail entity. For instance, the triple (\texttt{John}, \texttt{bornIn}, \texttt{USA}) states that \texttt{John} is born in the United States of America. 

While knowledge graphs can directly benefit a number of downstream tasks such as information retrieval~\cite{reinanda2020knowledge}, they can also be exploited to discover hidden patterns in a specific domain of application. For example, if in a knowledge graph, most people that are born in \texttt{Germany} and live in \texttt{Germany} are Germans, then \texttt{bornIn}(\texttt{x,Germany}) $\land$ \texttt{livesIn}(\texttt{x, Germany}) $\implies$ \texttt{isCitizenOf}(\texttt{x,\\Germany}) is a pattern that captures this observation. If in addition this is the case for most of the countries in the knowledge graph, then \texttt{type}(\texttt{Y,Country}) $\land$ \texttt{bornIn}(\texttt{x, Y}) $\land$ \texttt{livesIn}(\texttt{x,Y}) $\implies$ \texttt{isCitizenOf}(\texttt{x, Y}) generalizes the pattern described above to the relationships between people, their countries of birth, and their nationalities. 

Rule mining systems have been proposed as a means to automatically find hidden patterns in knowledge graphs~\cite{buczak2010fuzzy,dhar1993driven}. They are particularly useful for very large knowledge graphs with millions of facts where patterns cannot be induced by human experts due to the large amount of data to process. In recent years, numerous rule mining approaches have been developed \cite{GalarragaTHS13,galarraga2015fast,ijcai2019-435,ott2021safran}. These systems were shown to be more efficient in time complexity compared to inductive logic programming (ILP) approaches \cite{galarraga2015fast}. In particular, AMIE+ \cite{galarraga2015fast} appears to be one of the fastest rule mining systems to date. AnyBURL \cite{ijcai2019-435} is a bottom-up rule mining system tailored towards link prediction. It mines cyclic and acyclic rules through random walks in the input knowledge graph. The expressiveness of AnyBURL makes it generate thousands of rules within a short period of time. SAFRAN \cite{ott2021safran} systematically clusters rules generated by AnyBURL to accelerate inference during link prediction. Although these approaches achieve a remarkable performance on popular benchmark datasets such as WN18RR \cite{dettmers2018convolutional}, FB15K\cite{toutanova2015observed}, and YAGO3 \cite{mahdisoltani2014yago3}, their evaluation is often limited to runtime measurements, confidence scores of the generated rules, or link prediction tasks. In this work, we go beyond these evaluation settings and provide an in-depth analysis of the generated rules.

As opposed to rule mining systems, embedding-based methods first project a given knowledge graph (entities and relations) onto a low dimensional vector space and afterwards trained to model facts in the input graph via vector operations. In this process, entities and relations need not to be projected onto the same vector space. For instance, RESCAL \cite{nickel2011three} represents entities as vectors in $\mathbb{R}^d$, and relations as matrices in $\mathbb{R}^{d\times d}$ for some $d \in \mathbb{N}$. The latent representations of entities and relations are then used together with a scoring function to solve downstream tasks such as link prediction and knowledge graph completion~\cite{wang2017knowledge,dai2020survey}.

In this work, we investigate whether and how embedding-based knowledge completion on knowledge graphs can help discover new rules. In contrast to existing approaches that jointly learn rules and knowledge graph embeddings, we first compute the embeddings of a given knowledge graph using a lightweight embedding model. In a second step, we exploit the semantics embedded in the latent representation of entities and relations to infer new plausible triples that are added to the original knowledge graph. Finally, we apply a rule mining algorithm to the enriched knowledge graph. In our experiments, we find new rules that were previously out-of-reach. 

The rest of the paper is organized as follows: First, we present some related works on rule mining. Second, we introduce the notions needed for the rest of the paper. Next, we present our proposed approach in detail, and evaluate it on benchmark datasets. Finally, we discuss the results obtained, and conclude the paper.

\section{Related work}
There has been a significant amount of research in the application of knowledge graphs and link prediction to rule mining in various domains. In this section, we describe existing approaches for link prediction and rule mining on knowledge graphs. 


Inductive Logic Programming (ILP)~\cite{muggleton1991inductive} is a widely used approach to accomplish learning on multi relational data. However, as the size of the data grows, ILP-based systems become inherently slow and fail to compute quality solutions in most cases. Rule mining systems were developed to overcome the aforementioned limitations of ILP-based systems. AMIE~\cite{GalarragaTHS13} is a rule mining system specifically designed to support the open world assumption. To this end, the authors introduced a new confidence measure called \emph{partial completeness assumption} (PCA). Experiments on benchmark datasets showed that AMIE outperforms baseline approaches such as ALEPH\footnote{\url{http://www.cs.ox.ac.uk/activities/machlearn/Aleph/aleph_toc.html}} as it could generate high-precision rules while remaining scalable. Its recent version, AMIE+~\cite{galarraga2015fast}, applies additional optimization steps to prune the search space and reduce the runtime.
AnyBURL \cite{ijcai2019-435} mines cyclic and acyclic rules through random walks. Its expressiveness allows it to generate a large amount of rules within a short period of time. SAFRAN \cite{ott2021safran} systematically clusters rules generated by AnyBURL to accelerate inference during link prediction. Wang et al. \cite{wang1903logic} proposed a novel logic rule-enhanced method which can be integrated with any translation-based knowledge graph embedding model such as TransE. Guo et al. \cite{guo2016jointly} proposed to jointly embed knowledge graphs and logical rules to represent triples and rules in a unified framework. Their method was evaluated on link prediction and triple classification tasks. In another work,  the authors iteratively obtain soft rules (i.e. rules that  might not hold for a few exceptions) from data to guide knowledge graph embeddings~\cite{guo2018knowledge}. 
In IterE \cite{zhang2019iteratively}, knowledge graph embeddings are used to generate rules which are in turn used to infer missing information for sparse entities to further improve the embeddings. 
Ruleformer \cite{xu2022ruleformer} uses context information to address the problem of rule selection for inference tasks. Ruleformer consists of an encoder that extracts context information and a decoder that generates probabilities for target relations. The approach treats rule mining as a sequence-to-sequence task and uses a modified attention mechanism to parse input subgraphs. 
RuDik \cite{ahmadi2020mining} is a system that mines declarative rules over RDF knowledge graphs. It discovers both positive relationships between elements and negative patterns identifying data contradictions. The system increases the expressive power of the supported rule language and is robust to errors and missing data. In this sense, RuDiK is capable of identifying effective rules for different target applications while being computationally efficient. Islam et al. \cite{islam2022negative} proposed a method for learning low-dimensional vector representations of entities and relations to improve knowledge graph completion. The importance of efficient embeddings and explainable predictions is highlighted, and a new negative sampling method called \emph{Simple Negative Sampling} is proposed to address this. Additionally, a new rule mining method based on the learned embeddings is proposed to better understand the semantics of knowledge graphs. The overall system was evaluated on FB15K, FB15K-237, WN18, WN18RR, and YAGO3-10 and the results showed improved predictive performance and knowledge extraction capabilities. In another study, Ho et al. \cite{ho2018rule} proposed a rule mining method tailored towards missing fact prediction on incomplete knowledge graphs. The method iteratively extends rules induced from a knowlege graph by relying on the feedback from a precomputed embedding model and external information sources. This approach was shown to be effective on experiments with real-world datasets, producing high-quality learned rules and accurate fact predictions. Although rule mining is studied extensively in the literature, the effect of knowledge graph completion on rule mining via link prediction was not studied. Therefore, in this work, we use knowledge graph completion as the backbone of our novel rule mining approach and conduct extensive experiments to show its effectiveness. 

\section{Background}
In the rest of the paper, we denote a knowledge graph by $\kg$, its set of entities by $\ent$, and its set of relations by $\rel$. It is clear that $\kg \subseteq{\ent \times \rel \times \ent}$.
Unless otherwise specified, $\mathcal{V}$ is a countable set of variables.

\subsection{Knowledge graphs}
Knowledge graphs are collections of facts in the form of triples $(s, p, o)$, i.e., subject-predicate-object. For example, the triple (\texttt{Joe Biden}, \texttt{presidentOf}, \texttt{USA}) states that Joe Biden is the president of the United States of America. Knowledge graphs are useful for a number of downstream tasks, e.g., question answering \cite{hixon2015learning,lukovnikov2017neural}, information retrieval \cite{bounhas2020building,dietz2018utilizing}, and drug discovery \cite{bonner2021review,maclean2021knowledge}.

Knowledge graphs can also be projected onto low dimensional vector spaces to facilitate downstream tasks such as link prediction \cite{bordes2013translating}, triplet classification \cite{socher2013reasoning}, question answering \cite{bordes-etal-2014-question,bordes2014open}, and recommendation systems \cite{zhang2016collaborative,sun2018recurrent}. Prominent examples of knowledge graph embedding approaches include TransE \cite{bordes2013translating}, DistMult \cite{yang2015embedding}, RotatE \cite{sunrotate}, ComplEx \cite{trouillon2016complex}, and R-GCNs \cite{schlichtkrull2018modeling}. In this work, we employ the three approaches TransE, DistMult, and RotatE to learn embeddings, and leverage them to 1) enrich the input knowledge graph, and 2) mine Horn rules on the enriched knowledge graph.

\subsection{Atoms and rules}
Given a knowledge graph $\kg\subseteq{\ent \times \rel \times \ent}$, an expression of the form $r(X,Y)$, where $r\in \rel$ and $X, Y \in \ent \cup \mathcal{V}$ is called an atom. If $X\in \ent$ or $Y \in \ent$, then $r(X,Y)$ is said to be \textit{instantiated}. For example, \texttt{bornIn}(\texttt{x, France}) is an instantiated atom. If $X\in \ent$ and $Y\in \ent$, then $r(X,Y)$ is a fact which may or may not be in $\kg$.

Assume a set of atoms $A = \{A_i\}_{i=1}^{n}$. A conjunctive query of $A$ is an expression of the form $A_1\land A_2\ \land , \dots, \land\ A_n$.
Let $B=A_1\land A_2\ \land, \dots, \land\ A_n$ be a conjunctive query, and $H$ an atom. 
A rule with body $B$ and head $H$ is the written as $B \implies H$. A rule is Horn if all variables in its head also appear in its body. For instance, the first rule in the introduction, \texttt{bornIn}(\texttt{x,Germany}) $\land$ \texttt{livesIn}(\texttt{x, Germany}) $\implies$ \texttt{isCitizenOf}(\texttt{x,Germany}), is a Horn rule. In this work, we mine Horn rules using our proposed approach.

\subsection{Variable assignment and rule prediction}
Let $Q_{\kg}^{\mathcal{V}}$ be the set of all conjunctive queries that can be constructed from $\kg$ and $\mathcal{V}$. Note that a conjunctive query of length 1 is an atom.
A variable assignment $\sigma: Q_{\kg}^{\mathcal{V}} \longrightarrow Q_{\kg}^{\mathcal{V}}$ is a function that replaces all variables in a conjunctive query by constants in the knowledge graph $\kg$.

Let a rule $R = B \implies H$ such that $B=A_1\land A_2 \land, \dots, \land A_n$. A prediction of $R$ is a couple $(\sigma(B), \sigma(H))$ such that $\sigma(A_i) \in \kg\ \forall\ i=1, \dots, n$.
The prediction is said to be true if $\sigma(H)\in \kg$.

\subsection{Relation functionality}
Let $\kg\subseteq{\ent \times \rel \times \ent}$ be the reference knowledge graph.
The functionality score \cite{SuchanekAS11} of a relation $r\in\rel$ denoted by $fun(r)$ is defined as
\begin{eqnarray}
fun(r) = \frac{|\{s: \exists o: r(s,o)\in \kg\}|}{|\{(s,o): r(s,o)\in \kg\}|}.
\end{eqnarray}
The functionality score of the inverse relation $r^-$ is defined analogously:
\begin{eqnarray}
fun(r^-) = \frac{|\{o: \exists s: r(s,o)\in \kg\}|}{|\{(s,o): r(s,o)\in \kg\}|}.
\end{eqnarray}
Note that $fun(r) \in [0,1]$ and $fun(r^-) \in [0,1]$ for all relation $r\in \rel$.
The functionality score generalizes the notion of functional relations: a relation is functional if its functionality score is 1. This is equivalent to: a relation is functional if it couples a given subject with at most one object. Examples of functional relations include \texttt{birthDate}, and \texttt{birthPlace}.

\subsection{Support and confidence}
The support $\mathit{sup}$ of a rule $R = B \implies H$ is defined as the number of true predictions given by the rule in the reference knowledge graph:
\begin{eqnarray}
\mathit{support}(R) := |\{(\sigma(B), \sigma(H)): \sigma(H)\in \kg\}|.
\end{eqnarray}
The notion of confidence depends on the assumption that is used.
There are two common assumptions: the closed world assumption (CWA), and the open world assumption (OWA). Under the CWA, all facts that are not present in the reference knowledge graph are considered false. Consequently, a learning system under the CWA is constrained to reason only about facts present in the knowledge graph. On the other hand, the OWA assumes that facts that are not present in the reference knowledge graph are not necessarily false. This encourages the learning system to draw imprecise yet unjustifiable conclusions.
To alleviate the aforementioned issues, \cite{GalarragaTHS13} proposed the so called partial completeness assumption (PCA). The PCA is directly linked to the notion of relation functionality: If $r(s,o)\in \kg$ and $fun(r)\ge fun(r^-)$, then it is assumed that all predictions $r(s, o')$ that are not present in $\kg$ are false. If $fun(r) < fun(r^-)$, then all predictions $r(s', o)$ that are not observed in $\kg$ are false. A false prediction is called counter-example. We now define the confidence of a rule with respect to an assumption.

The confidence of a rule is the ratio of the number of true predictions to the number of true predictions and false predictions:
\begin{eqnarray}
ce = |\{(\sigma(B), \sigma(H)): \sigma(H)\in cex(R)\}|,\\
\mathit{confidence}(R) := \frac{\mathit{support}(R)}{\mathit{support}(R)+ce},
\end{eqnarray}
where $cex(R)$ is the set of counter-examples of $R$ w.r.t. the assumption. If the PCA is used, then the confidence is called PCA confidence. The standard confidence uses the CWA.


\section{Method: Improving rule mining via embedding-based link prediction}
In this section, we describe our approach to improve rule mining systems. Our approach can be summarized as follows: 1) compute embeddings of a given knowledge graph, 2) infer new links from the computed embeddings using a scoring function, 3) add the predicted links to the initial knowledge graph, and 4) apply the considered rule mining system on the enriched knowledge graph.

\subsection{Embedding  generation}
\label{embedding}
A given knowledge graph \kg is first embedded into a continuous vector space using one of the-state-of-the-art embedding models in the literature. In this work, we consider three embedding approaches TransE \cite{bordes2013translating}, DistMult \cite{yang2015embedding}, and RotatE \cite{sunrotate}, whose main components are described below.
In the following, $B$, $ B'$ are the batches of correct and corrupted triple embeddings, respectively. $\gamma>0$ is a margin, $\|.\|$ is the Euclidean norm, and $\odot$ is the element-wise multiplication (Hadamard product).

\subsubsection{TransE}
This is a lightweight and computationally efficient model for learning embeddings on large-scale knowledge graphs. It is also one of the first approaches proposed in this research area. TransE represents entities and relations as vectors in the same semantic space and is trained to minimize the margin ranking loss:
\begin{eqnarray*}
\sum_{(\mathbf{h},\mathbf{r},\mathbf{t})\in B}\sum_{(\mathbf{h}',\mathbf{r},\mathbf{t}') \in B'} \max(0, \gamma + \|\mathbf{h}+\mathbf{r}-\mathbf{t}\|- \|\mathbf{h}'+\mathbf{r}-\mathbf{t}'\|).
\end{eqnarray*} 

\subsubsection{DistMult}
DistMult is similar to TransE but its scoring function uses multiplications instead of additions. More precisely, the loss function for the DistMult model is defined as
\begin{eqnarray*}
\sum_{(\mathbf{h},\mathbf{r},\mathbf{t}) \in B}\sum_{(\mathbf{h}',\mathbf{r},\mathbf{t}')\in B'}\max(0, \gamma + \|\mathbf{h}\odot\mathbf{r}\odot\mathbf{t}\| -  \|\mathbf{h}'\odot\mathbf{r}\odot\mathbf{t}'\|).
\end{eqnarray*}
By replacing additions with multiplications, DistMult enjoys the same computational complexity as TransE while allowing for more expressiveness.
\subsubsection{RotatE}
This embedding approach is based on the intuition that symmetric, anti-symmetric, and inverse relations can be modeled by rotations in a multidimensional complex space. Its loss function is defined by:
\begin{eqnarray*}
-\log\sigma(\gamma - \|\mathbf{h}\circ\mathbf{r}-\mathbf{t}\|) - \sum_{i=1}^{n}\frac{1}{d}\log\sigma(\|\mathbf{h}_i'\circ\mathbf{r}-\mathbf{t}_i'\|-\gamma),
\end{eqnarray*}
where $\circ$ is the rotation operator, $d$ the dimension of the embedding space, $\sigma$ is the sigmoid function, and $n$ the number of negative samples per correct triple. The modulus of each component $\mathbf{r}_i\in \mathbb{C}$ of $\mathbf{r}$ is constrained to be $1$. In this way, each component corresponds to a counterclockwise rotation by some angle $\theta_i\in \mathbb{R}$. 
\subsection{Link prediction}
In this section, we assume that the given knowledge graph is incomplete, which is the case with most real world knowledge graphs, and we aim to find some of the missing triples. 
First, we define the scoring function $f$ that allows us to find plausible new facts to be added to the original knowledge graph. In accordance with the loss functions in Section \ref{embedding}, we define three scoring functions $f_{T}$, $f_{D}$, $f_{R}$ for TransE, DistMult and RotatE, respectively:
\begin{align}
\label{transe}
f_T(\mathbf{h}, \mathbf{r}, \mathbf{t}) &= sigmoid(-\|\mathbf{h}+\mathbf{r}-\mathbf{t}\|).\\
\label{distmult}
f_D(\mathbf{h}, \mathbf{r}, \mathbf{t}) &= sigmoid(-\|\mathbf{h}\odot\mathbf{r}\odot\mathbf{t}\|)\\
\label{rotate}
f_R(\mathbf{h}, \mathbf{r}, \mathbf{t}) &= sigmoid(-\|\mathbf{h}\circ\mathbf{r}-\mathbf{t}\|).
\end{align}
The function $f$ (referring to any of the functions defined above) is defined to produce a score between 0 and 1 for every input triple, and assigns high scores to correct triples. This allows us to sort candidate triples based on their scores and select a given number of them for KG completion. In the following, we describe our link prediction algorithm, which is applicable to any embedding model provided that we know its scoring function. A pseudo code of our approach is provided in Algorithms \ref{alg:find} and \ref{alg:infer}.
Given a knowledge graph \kg, we randomly generate arbitrary triples using relations and entities in \kg (Algorithm \ref{alg:find}, lines 1-to-5). The generation is carried out by sampling a user-specified maximum number of entities and relations from \kg. It is also possible to directly provide target relations to the algorithm. From the generated triples, we discard those that are already present in the knowledge graph (Algorithm \ref{alg:find}, lines 6-to-10). The top-$k$ ($k$ is a hyper-parameter) best scoring triples from the remaining ones are added to \kg, resulting in $\bar{\kg}$ (Algorithm \ref{alg:infer}). We call $\bar{\kg}$ the enriched knowledge graph of $\kg$.

\begin{algorithm}[ht]
	\caption{Function \textsc{FindCandidateTriples}}
	\label{alg:find}
	\textbf{Input}: A knowledge graph $\kg \subseteq{\ent \times \rel \times \ent}$, a set of target relations $\tgrel$\\
	\textbf{Hyper-parameters}: Number of entities $n$ (default 1000) and of relations $m$ (default 10) to be sampled. \\
	\textbf{Output}: Candidate triples $\mathcal{C}$
	\begin{algorithmic}[1]
	    \IF{$\tgrel$ is $\{\}$}
	    \STATE  $\tgrel \leftarrow \textsc{RandSample}(\rel, m)$ \quad \# \textit{Randomly and uniformly sample without repetition}
		\ENDIF
		\STATE $\ent_1 \leftarrow \textsc{RandSample}(\ent, n)$
		\STATE $\ent_2 \leftarrow \textsc{RandSample}(\ent, n) \text{ such that } \ent_1 \cap \ent_2 = \emptyset$
		\FOR{$r \in \tgrel$}
		\FOR{$h \in \ent_1$}
		\FOR{$t \in \ent_2$}
		\IF{($h, r, t$) $\not\in$ \kg}
		\STATE $\mathcal{C} \leftarrow \mathcal{C} \cup \{(h, r, t)\}$
		\ENDIF
		\ENDFOR
		\ENDFOR
		\ENDFOR
		\STATE \textbf{return} $\mathcal{C}$
	\end{algorithmic}
\end{algorithm}

\begin{algorithm}[tp]
	\caption{Function \textsc{InferNewTriples}}
	\label{alg:infer}
	\textbf{Input}: Knowledge graph (KG) $\kg$, candidate triples $\mathcal{C}$\\
	\textbf{Hyper-parameters}: Number top-$k$ of best scoring triples (default $50$)\\
	\textbf{Output}: Enriched KG $\bar{\kg}$
	\begin{algorithmic}[1]
		\STATE $\bar{\kg} \leftarrow \kg$
            \STATE Compute scores of all triples in $\mathcal{C}$ using $f$
            \STATE Find top-$k$ scores 
		\FOR{($h, r, t$) $\in$ $\mathcal{C}$}
        \IF{$f(\mathbf{h}, \mathbf{r}, \mathbf{t})$ in top-$k$ scores}
		\STATE $\bar{\kg} \leftarrow \bar{\kg} \cup \{(h, r, t)\}$
		\ENDIF
		\ENDFOR
		\STATE \textbf{return} $\bar{\kg}$
	\end{algorithmic}
\end{algorithm}
We use the notation $(h,r,t)$ to denote actual triples, and the bold font $(\mathbf{h},\mathbf{r},\mathbf{t})$ to denote their embeddings (vector representations).

\section{Experiments}

\subsection{Datasets}
To demonstrate the effectiveness of our approach, we conducted experiments on seven benchmark datasets. A description of each dataset is given in the next subsections.
\subsubsection{DRKG.}
The drug repurposing knowledge graph (DRKG) \cite{drkg2020} is a biological knowledge graph relating genes, compounds, diseases, biological processes, side effects and symptoms. It is constructed by aggregating information from DrugBank \cite{wishart2018drugbank}, Hetionet \cite{himmelstein2015heterogeneous}, GNBR \cite{percha2018global}, String \cite{szklarczyk2016string}, IntAct \cite{kerrien2012intact} and DGIdb \cite{freshour2021integration}, and data collected from recent publications particularly related to Covid19. DRKG contains 107 relations, 97,238 entities, and 5,874,261 triples.

\subsubsection{OPENBIOLINK.}
OPENBIOLINK \cite{10.1093/bioinformatics/btaa274} is a resource and evaluation framework for link prediction on heterogeneous biomedical graph data. It contains benchmark datasets as well as tools for creating custom benchmarks and evaluating models. In this work, we use the default dataset which is a direct and high-quality graph available on Zenedo~\footnote{\url{https://zenodo.org/record/3834052/files/HQ_DIR.zip?download=1}}. The dataset contains 28 relations, 184,635 entities, and 4,563,405 triples. Note that these statistics exclude the negative samples (triples) which also come with the original data. 

\subsubsection{WN18RR}
WN18RR \cite{dettmers2018convolutional} is an improved version of WN18 \cite{bordes2013translating} that does not contain inverse relations. The initial WN18 dataset was constructed from WordNet, and its test set mostly consisted of inverse relations from the training set. As a result, many rule-based systems could easily achieve remarkable results on the test set of WN18. WN18RR was proposed to alleviate this issue by ensuring that the prediction of triples in the test set requires the complete modeling of the whole graph. WN18RR comes with 11 relations, 40,943 entities, and 93,003 triples.

\subsubsection{CARCINOGENESIS}
The CARCINOGENESIS dataset was first made available in Prolog format by the Oxford University Machine Learning Group, and later converted into OWL format~\footnote{\url{https://dl-learner.org/community/carcinogenesis/}} using the DL-Learner parser. The dataset describes chemical compounds and their properties, categorized into classes. Additionally, all compounds can be classified into three main groups: 1) carcinogenic, 2) non-carcinogenic, or 3) unlabeled. The dataset has been extensively used for class expression learning in description logics \cite{Kouagou2022Learning, Heindorf2022EvoLearner, kouagou2023neural, lehmann2011class}. CARCINOGENESIS consists of 24 relations (including RDF(S) constructs such as \texttt{rdfs:subClassOf}, \texttt{rdf:type}, etc), 23,645 entities, and 96,939 triples.

\subsubsection{MUTAGENESIS}
The MUTAGENESIS~\footnote{\url{https://github.com/SmartDataAnalytics/DL-Learner/tree/develop/examples/mutagenesis}} dataset also describes chemical compounds and the relationships between them. In particular, it contains mutagens with different properties. The dataset is used as a benchmark for class expressions learning systems. It consists of 14 relations (including RDF(S) constructs), 15,260 entities, and 62,067 triples.

\subsubsection{FB15K-237}
FB15K-237 \cite{toutanova2015observed} is a subset of Freebase \cite{bollacker2008freebase}. It mainly describes facts about movies, actors and sports. The dataset was proposed as a replacement of the first version FB15K which contained several inverse relations in its test set, leading to questionable performance achieved by rule-based systems. FB15K-237 contains 237 relations, 14,541 entities, and 510,116 triples.

\subsubsection{YAGO3-10}
YAGO3-10 is benchmark dataset for knowledge base completion \cite{chami2020low,dettmers2018convolutional,zhang2020learning}. It is a subset of YAGO3 \cite{mahdisoltani2014yago3} where entities with less interactions (less than ten different relations with other entities) are removed. YAGO3-10 has 37 relations, 123,182 entities, and  1,089,040 triples. Most of its triples describe facts about persons such as profession, birth location, and gender.

\subsection{Experimental setup and hardware}
We trained each of the models TransE, DistMult, and RotatE for 200 epochs with an embedding dimension of 50 and a learning rate of 0.01. We used the PyKEEN~\cite{ali2021pykeen} library to train knowledge graph embeddings. The split into training, validation, and test sets for each dataset is specified in Table \ref{tab:statistics}. The training curves are shown in Figure \ref{fig:transe-train-curves} and the results achieved by each embedding model on the task of link prediction are presented in Table~\ref{tab:link-prediction}.
The trained embeddings are used to infer new triples, see Algorithms~\ref{alg:infer} and \ref{alg:find}, which are then added to the original graphs. The final step of our proposed approach applies a rule mining system on the augmented graphs to generate rich rules. The goal here is to identify and analyze new rules that could not be discovered in each original graph. Our proposed approach uses AMIE+ because of its scalability capabilities and the fact that its results are reproducible; AnyBURL~\cite{ijcai2019-435} generates different rules for each run.

We also conducted experiments on link prediction on the test sets using the generated rules. Here, we compare our approach with the recent state-of-the-art system AnyBURL~\cite{ijcai2019-435} in terms of \textit{Hits} scores (Table~\ref{tab:link-prediction-rule}) and runtimes (Figure~\ref{fig:runtime}).  All systems were run with their default settings on a server with 16 Intel Xeon E5-2695 CPUs @2.30GHz and 128GB RAM.

\subsection{Evaluation metrics}
We use two metrics, Hits@$k$ (with $k$ a positive integer) and the \emph{mean reciprocal rank} (MRR), to evaluate the quality of the computed embeddings and the mined rules on each knowledge graph. The two metrics are defined below.
\begin{align}
\label{metrics}
\text{Hits@}k =& \frac{1}{2|\kg_{test}|} \sum_{(e_1, r, e_2)\in \kg_{test}}\mathbbm{1}(\mathit{rank}[e_1|r,e_2]\le k) \nonumber\\ +& \mathbbm{1}(\mathit{rank}[e_2|e_1,r]\le k),\\\nonumber
\textsc{MRR} =& \frac{1}{2|\kg_{test}|} \sum_{(e_1, r, e_2)\in \kg_{test}}\frac{1}{\mathit{rank}[e_1|r,e_2]}\nonumber\\\nonumber
+& \frac{1}{\mathit{rank}[e_2|e_1,r]},
\end{align}
where $\kg_{test}$ is the test set, $\mathbbm{1}$ a unary function that takes a condition as input and returns 1 if the condition is satisfied, and 0 otherwise. Hits@$k$ measures the proportion of the correct head and tail entities that are ranked in the top$k$ predictions. MRR measures the average of the inverse ranks of the correct head and tail entities in the predictions. For both metrics, the higher the better.

\begin{table*}[th!]
\caption{Statistics of benchmark datasets}
\label{tab:statistics}
\begin{tabular}{|l|r|r|r|r|r|}
\hline
\multicolumn{1}{|c|}{Dataset} & \multicolumn{1}{c|}{Train Triples} & \multicolumn{1}{c|}{Valid. Triples} & \multicolumn{1}{c|}{Test. Triples} & \multicolumn{1}{c|}{\#Entities} & \multicolumn{1}{c|}{\#Relations} \\ \hline
WN18RR  & 78,401 & 4,189  & 10,413 & 40,943 & 11  \\ \hline
CARCINOGENESIS & 75,005 & 6,131  & 15,803  & 23,645 & 24  \\ \hline
MUTAGENESIS & 47,926 & 3,949 & 10,192 & 15,260 & 14\\ \hline
FB15K-237   & 223,871 & 24,606  & 261,639  & 14,541 & 237 \\ \hline
YAGO3-10  & 793,690 & 84,144  & 211,206 & 123,182 & 37  \\ \hline
DRKG & 4,239,782 & 466,825 & 1,167,654 & 97,238 & 107 \\ \hline
OPENBIOLINK & 3,334,890  & 350,669 & 877,846 & 184,635 & 28  \\ \hline
\end{tabular}
\end{table*}

\subsection{Results on link prediction}
\subsubsection{Link prediction with embedding models}
\label{link-pred-embedding}
\begin{figure*}[!th]
\centering
  \includegraphics[width=\textwidth]{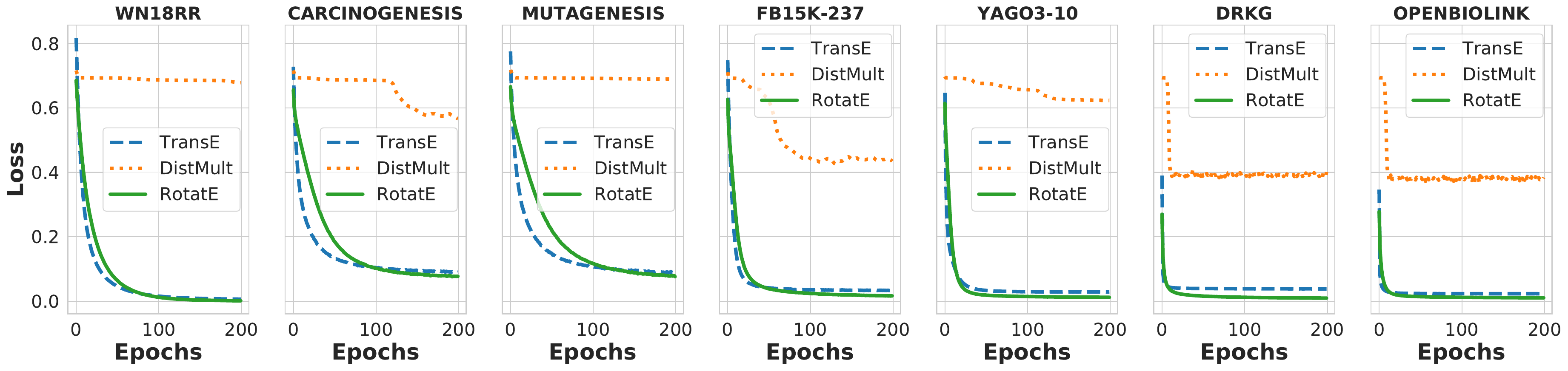}
\caption{Loss during embedding computation}
\label{fig:transe-train-curves}
\end{figure*}

Each knowledge graph was split into training and test sets of triples such that there are no out-of vocabulary entities and relations at inference time. The embedding models TransE, DistMult, and RotatE were then trained on the training set for 200 epochs.
Figure \ref{fig:transe-train-curves} shows the progression of the loss function during embedding computation on each dataset. From the figure, we can observe that the loss functions of TransE and RotatE decrease on all datasets, and approach 0 in most cases. This suggests that the two models are able to learn embeddings that effectively model the input knowledge graphs. The loss functions of DistMult also decrease in most cases but stagnate early and remain far from those of TransE and RotatE. As a result, the embeddings computed by DistMult are of low quality as can be seen in Table~\ref{tab:link-prediction}. Although one might achieve better results via hyperparameter search, we chose to use the default parameters in PyKEEN~\cite{ali2021pykeen}. It is also worth noting that by using recent state-of-the-art embedding models in the literature, one might achieve better results in the experiments we conduct herein.

Once training is completed, we use entity and relation embeddings as well as scoring functions (see Equations~\ref{transe}, \ref{distmult}, and \ref{rotate}) for each model to compute the likelihood of each triple in the test set. The latter are used to rank target head and tail entities among all other entities and compute the metrics Hits@k and \textit{mean reciprocal rank} (MRR) defined in Equation \ref{metrics}. Tabe~\ref{tab:link-prediction} presents the results achieved by each model on different datasets.
We can observe that RoatE achieves the highest performance on all datasets with up to 43\% Hits@1 and 47\% MRR on WN18RR. Overall, TransE achieves the second best performance except for Hits@1 on WN18RR where it is outperformed by DistMult.
The poor performance of DistMult is justified by its inability to train properly with the default hyper-parameters, see for example high loss values in Figure~\ref{fig:transe-train-curves} for this model. The best entity and relation embeddings are therefore computed by RotatE. We use the latter in our approach to generate and analyze rules on DRKG in Section~\ref{sec:analyze-drkg}.
\begin{table}[h!]
\caption{Link prediction results on the test sets using embedding models}
\label{tab:link-prediction}
\centering
\resizebox{0.48\textwidth}{!}{
\begin{tabular}{|l|r|r|r|r|}
 \hline 
\multicolumn{1}{|c|}{Dataset} & \multicolumn{1}{c|}{Hits@1} & \multicolumn{1}{c|}{Hits@3} & \multicolumn{1}{c|}{Hits@10} & \multicolumn{1}{c|}{MRR} \\
\hline 
\multicolumn{1}{|c}{} & \multicolumn{4}{c|}{TransE} \\ 
\hline 
WN18RR & 0.02 & 0.35 & 0.49 & 0.20 \\ 
\hline 
CARCINOGENESIS & 0.13 & 0.22 & 0.32 & 0.19 \\ 
\hline 
MUTAGENESIS & 0.09 & 0.20 & 0.31 & 0.16 \\ 
\hline 
FB15K-237 & 0.13 & 0.26 & 0.42 & 0.23 \\ 
\hline 
YAGO3-10 & 0.06 & 0.11 & 0.20 & 0.11 \\ 
\hline 
DRKG & 0.01 & 0.04 & 0.12 & 0.05 \\ 
\hline 
OPENBIOLINK & 0.02 & 0.06 & 0.18 & 0.07 \\ 
\hline 
\multicolumn{1}{|c}{} & \multicolumn{4}{c|}{DistMult} \\ 
\hline 
WN18RR & 0.03 & 0.03 & 0.03 & 0.03 \\ 
\hline 
CARCINOGENESIS & 0.09 & 0.14 & 0.22 & 0.14 \\ 
\hline 
MUTAGENESIS & 0.00 & 0.00 & 0.01 & 0.00 \\ 
\hline 
FB15K-237 & 0.05 & 0.10 & 0.21 & 0.10 \\ 
\hline 
YAGO3-10 & 0.01 & 0.02 & 0.03 & 0.02 \\ 
\hline 
DRKG & 0.00 & 0.01 & 0.02 & 0.01 \\ 
\hline 
OPENBIOLINK & 0.00 & 0.01 & 0.01 & 0.01 \\ 
\hline 
\multicolumn{1}{|c}{} & \multicolumn{4}{c|}{RotatE} \\ 
\hline 
WN18RR & 0.43 & 0.49 & 0.54 & 0.47 \\ 
\hline 
CARCINOGENESIS & 0.26 & 0.37 & 0.48 & 0.33 \\ 
\hline 
MUTAGENESIS & 0.28 & 0.39 & 0.51 & 0.36 \\ 
\hline 
FB15K-237 & 0.17 & 0.31 & 0.49 & 0.27 \\ 
\hline 
YAGO3-10 & 0.07 & 0.13 & 0.25 & 0.13 \\ 
\hline 
DRKG & 0.04 & 0.12 & 0.28 & 0.11 \\ 
\hline 
OPENBIOLINK & 0.05 & 0.11 & 0.27 & 0.12 \\ 
\hline 
\end{tabular}}
\end{table}

\subsubsection{Link prediction with rule mining systems}
First, we report the numbers of rules generated by each rule mining system in Table~\ref{tab:mined-rules}. AnyBURL was set to run for 500 seconds on each dataset while AMIE+ and our system ran without interruption. In fact, AnyBURL was designed to generate a large number of rules within a short period of time; this is not the case for the other two systems. From Table~\ref{tab:mined-rules}, we can observe that AnyBURL mines approximately 500$\times$ more rules than AMIE+ and our system.
\begin{table}[tbh!]
\caption{Number of rules generated by each approach}
\label{tab:mined-rules}
\resizebox{0.48\textwidth}{!}{
\begin{tabular}{|l|r|r|r|r|r|}
 \hline 
\multicolumn{1}{|c|}{Dataset} & \multicolumn{1}{c|}{AMIE+} & \multicolumn{1}{c|}{Ours} & \multicolumn{1}{c|}{AnyBURL}\\ 
 \hline 
WN18RR  & 34 & 33 & 35,164\\ 
\hline 
CARCINOGENESIS  & 185 & 170 & 177,924\\ 
\hline 
MUTAGENESIS  & 9 & 9 & 269,660\\ 
\hline 
FB15K-237  & 7,591 & 5,609 & 1,414,568\\ 
\hline 
YAGO3-10  & 235 & 185 & 852,324\\ 
\hline 
DRKG  & 2,804 & 2,745 & 1,573,497\\ 
\hline 
OPENBIOLINK  & 713 & 607 & 402,256\\ 
\hline 
\end{tabular}}
\end{table}

\begin{figure*}[!th]
\centering
  \includegraphics[width=\textwidth]{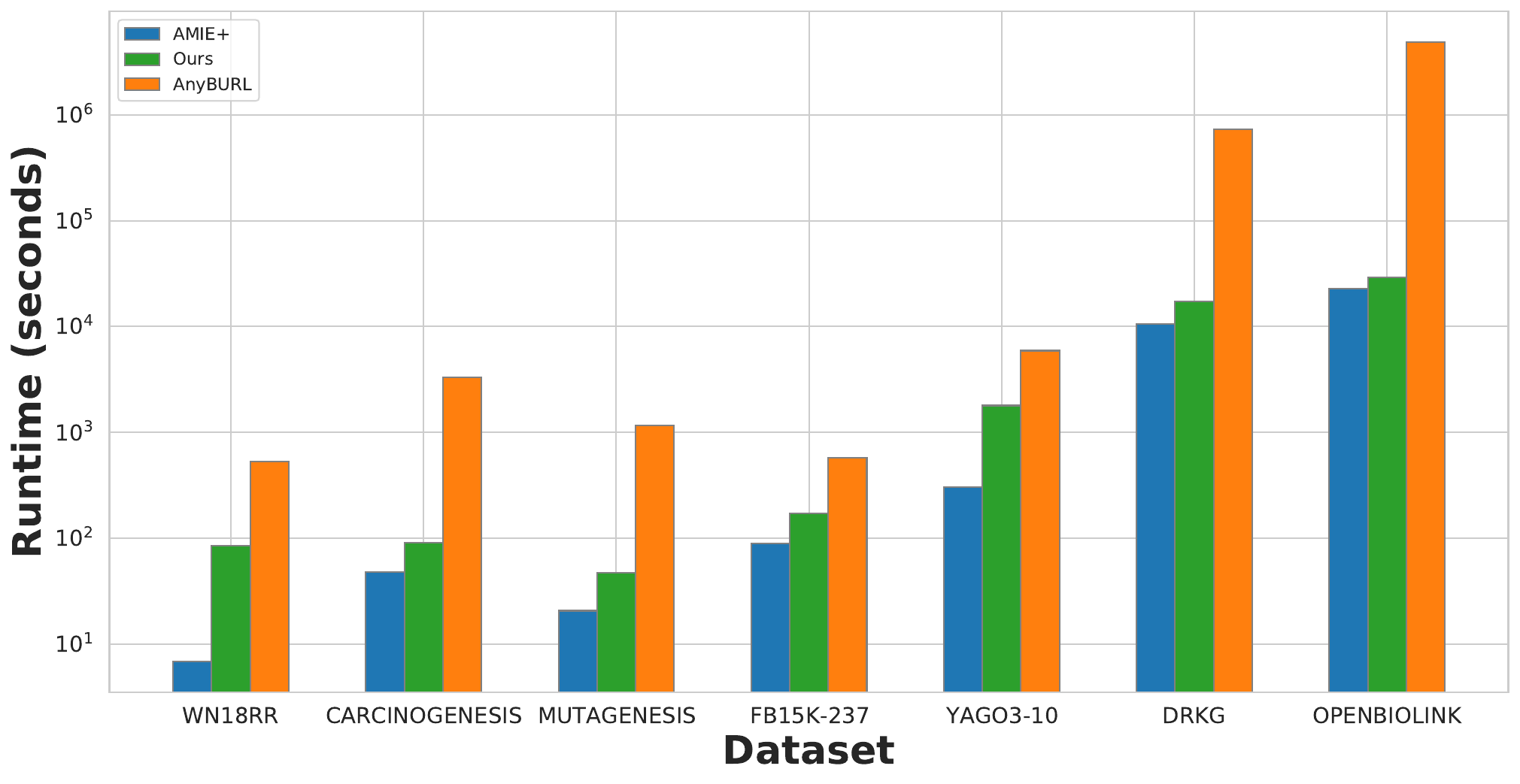}
\caption{Total runtime of each rule mining system. The y-axis is log-scaled for a better visualization due to large differences in runtime across datasets and approaches}
\label{fig:runtime}
\end{figure*}

We now compare the three systems in terms of total execution time on each dataset. For AMIE+ and AnyBURL, we consider rule mining and rule application runtimes. In addition to these two runtimes, we consider the embedding computation time for our approach. Figure \ref{fig:runtime} depicts the total runtime in a logarithmic scale for all three approaches. Although AnyBURL generates a large number of rules in a short period of time, the application of the latter to solve link prediction tasks remains inevitably time intensive. As a result, AnyBURL took nearly (over) a million seconds to complete on OPENBIOLINK (DRKG), as can be seen in the figure. As expected, AMIE+ is slightly faster than our proposed approach due to the fact that the latter initially requires embedding computation. Nonetheless, our approach remains overall faster than AnyBURL as it generates a smaller and concise set of rules hence allowing for their efficient application.

We use the mined rules to perform predictions of head and tail entities in the test sets. We use the same training, validation and test-splits as in Section~\ref{link-pred-embedding}. Table~\ref{tab:link-prediction-rule} presents the results. On 5 out of 7 datasets, the expressive rule mining system AnyBURL outperforms AMIE+ and our approach (using the best embedding model RotatE). On the two other datasets which are the largest, we observe the opposite. Regarding the least expressive approaches AMIE+ and ours, we observe comparable performance on all datasets. This suggests that in addition to being highly scalable, lightweight rule mining systems (e.g., ours) might be a better option on large datasets in terms of quality of the mined rules.

\begin{table}[h!]
\caption{Link prediction results on the test sets using rule mining systems}
\label{tab:link-prediction-rule}
\centering
\resizebox{0.48\textwidth}{!}{
\begin{tabular}{|l|r|r|r|}
 \hline 
 \multicolumn{1}{|c|}{Dataset} & \multicolumn{1}{c|}{Hits@1} & \multicolumn{1}{c|}{Hits@3} & \multicolumn{1}{c|}{MRR} \\ 
 \hline 
\multicolumn{1}{|c}{} & \multicolumn{3}{c|}{AMIE+} \\ 
\hline 
WN18RR & 0.390 & 0.395 & 0.398 \\ 
\hline 
CARCINOGENESIS & 0.216 & 0.245 & 0.248 \\ 
\hline 
MUTAGENESIS & 0.217 & 0.244 & 0.244 \\ 
\hline 
FB15K-237 & 0.180 & 0.284 & 0.389 \\ 
\hline 
YAGO3-10 & 0.244 & 0.333 & 0.408 \\ 
\hline 
DRKG & 0.147 & 0.172 & 0.215 \\ 
\hline 
OPENBIOLINK & 0.038 & 0.081 & 0.158 \\ 
\hline 
\multicolumn{1}{|c}{} & \multicolumn{3}{c|}{Ours (with RotatE)} \\ 
\hline 
WN18RR & 0.384 & 0.390 & 0.392 \\ 
\hline 
CARCINOGENESIS & 0.220 & 0.248 & 0.252 \\ 
\hline 
MUTAGENESIS & 0.182 & 0.215 & 0.227 \\ 
\hline 
FB15K-237 & 0.177 & 0.280 & 0.385 \\ 
\hline 
YAGO3-10 & 0.243 & 0.331 & 0.405 \\ 
\hline 
DRKG & 0.147 & 0.172 & 0.215 \\ 
\hline 
OPENBIOLINK & 0.059 & 0.129 & 0.258 \\ 
\hline 
\multicolumn{1}{|c}{} & \multicolumn{3}{c|}{AnyBURL} \\ 
\hline 
WN18RR & 0.469 & 0.527 & 0.594 \\ 
\hline 
CARCINOGENESIS & 0.483 & 0.533 & 0.630 \\ 
\hline 
MUTAGENESIS & 0.477 & 0.547 & 0.639 \\ 
\hline 
FB15K-237 & 0.278 & 0.424 & 0.569 \\ 
\hline 
YAGO3-10 & 0.372 & 0.457 & 0.531 \\ 
\hline 
DRKG & 0.064 & 0.075 & 0.090 \\ 
\hline 
OPENBIOLINK & 0.009 & 0.016 & 0.026 \\ 
\hline
\end{tabular}}
\end{table}

\subsection{Main results}
Because our system is built on top of AMIE+, we conduct experiments to quantify the impact of our contribution. We systematically analyze rules generated by the two approaches. Recall our main contribution is concerned with knowledge graph enrichment: Given a knowledge graph, we train an embedding model on it and use the learned embeddings to infer new facts. These facts are added to the original graph (this leads to an enriched graph $\bar{\kg}$), and we apply AMIE+ on the enriched graph. We refer to rules generated on the enriched graph as \emph{rules after enrichment}, and to those generated on the original graph as \emph{rules before enrichment}. Our hypothesis was that the two groups of rules would overlap, but also that we would discover new rules after enrichment. The results of our experiments\textemdash see Tables \ref{tab:transe-generated}, \ref{tab:distmult-generated}, and \ref{tab:rotate-generated}\textemdash clearly undergird this hypothesis. These tables give the counts of the generated rules for TransE, DistMult, and RotatE, respectively. Note the limited amount of rules discovered on small datasets such as MUTAGENESIS and WN18RR. On such datasets, the number of new rules discovered after enrichment is proportionally low; for instance, no new rules were found on MUTAGENESIS. 

Tables~\ref{tab:transe-confidence}, \ref{tab:distmult-confidence}, and \ref{tab:rotate-confidence} present the standard (Std) and partial completeness  assumption (PCA) confidence scores of the generated rules categorized into the following five groups.
\begin{enumerate}
\item \emph{Confidence scores before knowledge graph enrichment.} We measure the average confidence scores of rules before the enrichment of each knowledge graph. The scores are calculated on each original knowledge graph.

\item \emph{Confidence scores after knowledge graph enrichment.} We generate and compute the standard and partial completeness assumption confidence scores of rules on each enriched knowledge graph $\bar{\kg}$.

\item \emph{Confidence scores of new rules.} These scores are extracted from the enriched knowledge graph $\bar{\kg}$. In other words, they represent the confidence scores of new rules generated on $\bar{\kg}$. 

\item \emph{Confidence scores of common (same) rules.} Some rules generated using the original knowledge graph are also found in the set of rules generated using the enriched knowledge graph $\bar{\kg}$. We refer to these rules as \emph{same rules}. We measure their average confidence scores on the original knowledge graph. 

\item \emph{Confidence scores of dropped rules.} We refer to rules that are generated using the original knowledge graph but that could not be retrieved when using \kg as \emph{dropped rules}. Their confidence scores are calculated on the original knowledge graph.
\end{enumerate}
In terms of confidence scores, our approach generates rules that are similar to those generated by AMIE+. However, we are able to discover new rules that could not be found by AMIE+. In addition, our approach discards (i.e., does not generate) low confidence rules that are generated by AMIE+ on the original graphs. This suggests that not only our approach can find new valuable rules, but it can also help cure those generated by AMIE+.
\begin{table}[h!]
\caption{Counts of rules generated on each knowledge graph using the TransE model. Here top-$k$ refers to the number of best scoring triples added to the original graph to form the enriched graph $\bar{\kg}$, see Algorithm~\ref{alg:infer}}
\label{tab:transe-generated}
\resizebox{0.48\textwidth}{!}{
\begin{tabular}{|l|r|r|r|r|r|}
 \hline 
\multicolumn{6}{|c|}{top-$k$ = 50} \\ 
\hline 
\multicolumn{1}{|c|}{Dataset} & \multicolumn{1}{c|}{\#Before} & \multicolumn{1}{c|}{\#After} & \multicolumn{1}{c|}{\#New} & \multicolumn{1}{c|}{\#Dropped} & \multicolumn{1}{c|}{\#Same} \\ 
 \hline 
WN18RR  & 34 & 33 & 7 & 8 & 26\\ 
\hline 
CARCINOGENESIS  & 185 & 170 & 23 & 38 & 147\\ 
\hline 
MUTAGENESIS  & 9 & 9 & 0 & 0 & 9\\ 
\hline 
FB15K-237  & 7,591 & 5,609 & 467 & 2,449 & 5,142\\ 
\hline 
YAGO3-10  & 235 & 185 & 5 & 55 & 180\\ 
\hline 
DRKG  & 2,804 & 2,745 & 1,433 & 1,492 & 1,312\\ 
\hline 
OPENBIOLINK  & 713 & 607 & 32 & 138 & 575\\ 
\hline 
\multicolumn{6}{|c|}{top-$k$ = 500} \\ 
\hline 
WN18RR  & 34 & 76 & 51 & 9 & 25\\ 
\hline 
CARCINOGENESIS  & 185 & 171 & 24 & 38 & 147\\ 
\hline 
MUTAGENESIS  & 9 & 9 & 0 & 0 & 9\\ 
\hline 
FB15K-237  & 7,591 & 5,751 & 616 & 2,456 & 5,135\\ 
\hline 
YAGO3-10  & 235 & 184 & 5 & 56 & 179\\ 
\hline 
DRKG  & 2804 & 2,744 & 1,434 & 1,494 & 1,310\\ 
\hline 
OPENBIOLINK  & 713 & 612 & 37 & 138 & 575\\ 
\hline 
\multicolumn{6}{|c|}{top-$k$ = 5000} \\ 
\hline 
WN18RR  & 34 & 152 & 123 & 5 & 29\\ 
\hline 
CARCINOGENESIS  & 185 & 228 & 81 & 38 & 147\\ 
\hline 
MUTAGENESIS  & 9 & 9 & 0 & 0 & 9\\ 
\hline 
FB15K-237  & 7,591 & 5,937 & 858 & 2,512 & 5,079\\ 
\hline 
YAGO3-10  & 235 & 177 & 5 & 63 & 172\\ 
\hline 
DRKG  & 2,804 & 2,743 & 1434 & 1,495 & 1,309\\ 
\hline 
OPENBIOLINK  & 713 & 613 & 41 & 141 & 572\\ 
\hline 
\end{tabular}}
\end{table}
\begin{table}[h!]
\caption{Counts of rules generated on each knowledge graph using the DistMult model}
\label{tab:distmult-generated}
\resizebox{0.48\textwidth}{!}{
\begin{tabular}{|l|r|r|r|r|r|}
 \hline 
\multicolumn{6}{|c|}{top-$k$ = 50} \\ 
\hline 
\multicolumn{1}{|c|}{Dataset} & \multicolumn{1}{c|}{\#Before} & \multicolumn{1}{c|}{\#After} & \multicolumn{1}{c|}{\#New} & \multicolumn{1}{c|}{\#Dropped} & \multicolumn{1}{c|}{\#Same} \\ 
 \hline 
WN18RR  & 34 & 28 & 2 & 8 & 26\\ 
\hline 
CARCINOGENESIS  & 185 & 170 & 23 & 38 & 147\\ 
\hline 
MUTAGENESIS  & 9 & 13 & 4 & 0 & 9\\ 
\hline 
FB15K-237  & 7,591 & 5,500 & 345 & 2,436 & 5,155\\ 
\hline 
YAGO3-10  & 235 & 185 & 5 & 55 & 180\\ 
\hline 
DRKG  & 2,804 & 2,745 & 1,433 & 1,492 & 1,312\\ 
\hline 
OPENBIOLINK  & 713 & 608 & 33 & 138 & 575\\ 
\hline 
\multicolumn{6}{|c|}{top-$k$ = 500} \\ 
\hline 
WN18RR  & 34 & 27 & 2 & 9 & 25\\ 
\hline 
CARCINOGENESIS  & 185 & 172 & 27 & 40 & 145\\ 
\hline 
MUTAGENESIS  & 9 & 9 & 0 & 0 & 9\\ 
\hline 
FB15K-237  & 7,591 & 5,513 & 352 & 2,430 & 5,161\\ 
\hline 
YAGO3-10  & 235 & 185 & 5 & 55 & 180\\ 
\hline 
DRKG  & 2,804 & 2,745 & 1,433 & 1,492 & 1,312\\ 
\hline 
OPENBIOLINK  & 713 & 605 & 33 & 141 & 572\\ 
\hline 
\multicolumn{6}{|c|}{top-$k$ = 5000} \\ 
\hline 
WN18RR  & 34 & 24 & 1 & 11 & 23\\ 
\hline 
CARCINOGENESIS  & 185 & 169 & 33 & 49 & 136\\ 
\hline 
MUTAGENESIS  & 9 & 15 & 6 & 0 & 9\\ 
\hline 
FB15K-237  & 7,591 & 5,406 & 382 & 2,567 & 5,024\\ 
\hline 
YAGO3-10  & 235 & 187 & 7 & 55 & 180\\ 
\hline 
DRKG  & 2,804 & 2,788 & 1,480 & 1,496 & 1,308\\ 
\hline 
OPENBIOLINK  & 713 & 609 & 60 & 164 & 549\\ 
\hline 
\end{tabular}}
\end{table}


\begin{table}[h!]
\caption{Counts of rules generated on each knowledge graph using the RotatE model}
\label{tab:rotate-generated}
\resizebox{0.48\textwidth}{!}{
\begin{tabular}{|l|r|r|r|r|r|}
 \hline 
\multicolumn{6}{|c|}{top-$k$ = 50} \\ 
\hline 
\multicolumn{1}{|c|}{Dataset} & \multicolumn{1}{c|}{\#Before} & \multicolumn{1}{c|}{\#After} & \multicolumn{1}{c|}{\#New} & \multicolumn{1}{c|}{\#Dropped} & \multicolumn{1}{c|}{\#Same} \\ 
 \hline 
WN18RR  & 34 & 28 & 2 & 8 & 26\\ 
\hline 
CARCINOGENESIS  & 185 & 170 & 23 & 38 & 147\\ 
\hline 
MUTAGENESIS  & 9 & 9 & 0 & 0 & 9\\ 
\hline 
FB15K-237  & 7,591 & 5,576 & 425 & 2,440 & 5,151\\ 
\hline 
YAGO3-10  & 235 & 185 & 5 & 55 & 180\\ 
\hline 
DRKG  & 2,804 & 2,745 & 1,433 & 1492 & 1,312\\ 
\hline 
OPENBIOLINK  & 713 & 609 & 36 & 140 & 573\\ 
\hline 
\multicolumn{6}{|c|}{top-$k$ = 500} \\ 
\hline 
WN18RR  & 34 & 28 & 2 & 8 & 26\\ 
\hline 
CARCINOGENESIS  & 185 & 203 & 54 & 36 & 149\\ 
\hline 
MUTAGENESIS  & 9 & 9 & 0 & 0 & 9\\ 
\hline 
FB15K-237  & 7,591 & 5,738 & 602 & 2,455 & 5,136\\ 
\hline 
YAGO3-10  & 235 & 184 & 7 & 58 & 177\\ 
\hline 
DRKG  & 2,804 & 2,747 & 1,437 & 1,494 & 1,310\\ 
\hline 
OPENBIOLINK  & 713 & 607 & 43 & 149 & 564\\ 
\hline 
\multicolumn{6}{|c|}{top-$k$ = 5000} \\ 
\hline 
WN18RR  & 34 & 390 & 367 & 11 & 23\\ 
\hline 
CARCINOGENESIS  & 185 & 207 & 71 & 49 & 136\\ 
\hline 
MUTAGENESIS  & 9 & 9 & 0 & 0 & 9\\ 
\hline 
FB15K-237  & 7,591 & 5,890 & 807 & 2,508 & 5,083\\ 
\hline 
YAGO3-10  & 235 & 198 & 26 & 63 & 172\\ 
\hline 
DRKG  & 2,804 & 2,998 & 1,690 & 1,496 & 1,308\\ 
\hline 
OPENBIOLINK  & 713 & 587 & 52 & 178 & 535\\ 
\hline 
\end{tabular}}
\end{table}

\begin{table*}[h!]
\caption{Average confidence scores using the TransE embedding model}
\label{tab:transe-confidence}
\resizebox{\textwidth}{!}{
\begin{tabular}{|l|cc|cc|cc|cc|cc|}
 \hline \multicolumn{11}{|c|}{top-$k$ = 50} \\ \hline 
 \multirow{2}{*}{Dataset} & \multicolumn{2}{c|}{\textbf{Rules before}} & \multicolumn{2}{c|}{\textbf{Rules after}} & \multicolumn{2}{c|}{\textbf{New rules}}    & \multicolumn{2}{c|}{\textbf{Dropped rules}}& \multicolumn{2}{c|}{\textbf{Same rules}} \\ \cline{2-11} & \multicolumn{1}{p{1.5cm}}{\centering Std.\ conf.}    & \multicolumn{1}{p{1.5cm}|} {\centering PCA \ conf. }  & \multicolumn{1}{p{1.5cm}}{\centering Std. \ conf.}& \multicolumn{1}{p{1.5cm}|}{\centering PCA \ conf.}  & \multicolumn{1}{p{1.5cm}}{\centering Std.\  conf.} & \multicolumn{1}{p{1.5cm}|} {\centering PCA \  conf.}  & \multicolumn{1}{p{1.5cm}}{\centering Std. \  conf.} &\multicolumn{1}{p{1.5cm}|}{\centering PCA \ conf.} & \multicolumn{1}{p{1.5cm}}{\centering Std. \  conf.} & \multicolumn{1}{p{1.5cm}|}{\centering PCA \ conf.} \\ 
\hline
WN18RR & 0.25 & 0.49 & 0.26 & 0.52 & 0.42 & 0.70 & 0.19 & 0.40 & 0.26 & 0.51 \\ 
\hline 
CARCINOGENESIS & 0.52 & 0.74 & 0.49 & 0.71 & 0.54 & 0.73 & 0.53 & 0.75 & 0.51 & 0.73 \\ 
\hline 
MUTAGENESIS & 0.44 & 0.54 & 0.47 & 0.54 & - & - & - & - & 0.44 & 0.54 \\ 
\hline 
FB15K-237 & 0.31 & 0.56 & 0.29 & 0.56 & 0.22 & 0.41 & 0.22 & 0.51 & 0.35 & 0.59 \\ 
\hline 
YAGO3-10 & 0.19 & 0.32 & 0.15 & 0.31 & 0.22 & 0.27 & 0.16 & 0.26 & 0.20 & 0.34 \\ 
\hline 
DRKG & 0.17 & 0.29 & 0.17 & 0.28 & 0.14 & 0.26 & 0.14 & 0.27 & 0.20 & 0.31 \\ 
\hline 
OPENBIOLINK & 0.28 & 0.32 & 0.24 & 0.27 & 0.46 & 0.51 & 0.27 & 0.30 & 0.28 & 0.32 \\ 
\hline 
\multicolumn{11}{|c|}{top-$k$ = 500} \\ 
\hline 
WN18RR & 0.25 & 0.49 & 0.51 & 0.74 & 0.65 & 0.86 & 0.17 & 0.37 & 0.27 & 0.53 \\ 
\hline 
CARCINOGENESIS & 0.52 & 0.74 & 0.49 & 0.71 & 0.52 & 0.71 & 0.53 & 0.75 & 0.51 & 0.73 \\ 
\hline 
MUTAGENESIS & 0.44 & 0.54 & 0.46 & 0.53 & - & - & - & - & 0.44 & 0.54 \\ 
\hline 
FB15K-237 & 0.31 & 0.56 & 0.30 & 0.56 & 0.27 & 0.47 & 0.22 & 0.51 & 0.35 & 0.59 \\ 
\hline 
YAGO3-10 & 0.19 & 0.32 & 0.15 & 0.31 & 0.22 & 0.27 & 0.15 & 0.26 & 0.20 & 0.34 \\ 
\hline 
DRKG & 0.17 & 0.29 & 0.17 & 0.28 & 0.14 & 0.26 & 0.14 & 0.27 & 0.20 & 0.31 \\ 
\hline 
OPENBIOLINK & 0.28 & 0.32 & 0.24 & 0.27 & 0.43 & 0.48 & 0.27 & 0.30 & 0.28 & 0.32 \\ 
\hline 
\multicolumn{11}{|c|}{top-$k$ = 5000} \\ 
\hline 
WN18RR & 0.25 & 0.49 & 0.47 & 0.68 & 0.53 & 0.72 & 0.15 & 0.20 & 0.26 & 0.54 \\ 
\hline 
CARCINOGENESIS & 0.52 & 0.74 & 0.46 & 0.70 & 0.41 & 0.67 & 0.53 & 0.75 & 0.51 & 0.73 \\ 
\hline 
MUTAGENESIS & 0.44 & 0.54 & 0.39 & 0.44 & - & - & - & - & 0.44 & 0.54 \\ 
\hline 
FB15K-237 & 0.31 & 0.56 & 0.29 & 0.56 & 0.24 & 0.49 & 0.22 & 0.50 & 0.35 & 0.59 \\ 
\hline 
YAGO3-10 & 0.19 & 0.32 & 0.15 & 0.31 & 0.21 & 0.27 & 0.15 & 0.25 & 0.21 & 0.35 \\ 
\hline 
DRKG & 0.17 & 0.29 & 0.17 & 0.28 & 0.14 & 0.26 & 0.14 & 0.27 & 0.20 & 0.31 \\ 
\hline 
OPENBIOLINK & 0.28 & 0.32 & 0.24 & 0.27 & 0.41 & 0.48 & 0.26 & 0.30 & 0.28 & 0.32 \\ 
\hline 
\end{tabular}}
\end{table*}

\begin{table*}[h!]
\caption{Average confidence scores using the DistMult model}
\label{tab:distmult-confidence}
\resizebox{\textwidth}{!}{\begin{tabular}{|l|cc|cc|cc|cc|cc|}
 \hline \multicolumn{11}{|c|}{top-$k$ = 50} \\ \hline 
 \multirow{2}{*}{Dataset} & \multicolumn{2}{c|}{\textbf{Rules before}} & \multicolumn{2}{c|}{\textbf{Rules after}} & \multicolumn{2}{c|}{\textbf{New rules}}    & \multicolumn{2}{c|}{\textbf{Dropped rules}}& \multicolumn{2}{c|}{\textbf{Same rules}} \\ \cline{2-11} & \multicolumn{1}{p{1.5cm}}{\centering Std.\ conf.}    & \multicolumn{1}{p{1.5cm}|} {\centering PCA \ conf. }  & \multicolumn{1}{p{1.5cm}}{\centering Std. \ conf.}& \multicolumn{1}{p{1.5cm}|}{\centering PCA \ conf.}  & \multicolumn{1}{p{1.5cm}}{\centering Std.\  conf.} & \multicolumn{1}{p{1.5cm}|} {\centering PCA \  conf.}  & \multicolumn{1}{p{1.5cm}}{\centering Std. \  conf.} &\multicolumn{1}{p{1.5cm}|}{\centering PCA \ conf.} & \multicolumn{1}{p{1.5cm}}{\centering Std. \  conf.} & \multicolumn{1}{p{1.5cm}|}{\centering PCA \ conf.} \\ 
\hline
WN18RR & 0.25 & 0.49 & 0.27 & 0.50 & 0.95 & 1.00 & 0.19 & 0.40 & 0.26 & 0.51 \\ 
\hline 
CARCINOGENESIS & 0.52 & 0.74 & 0.49 & 0.71 & 0.54 & 0.73 & 0.53 & 0.75 & 0.51 & 0.73 \\ 
\hline 
MUTAGENESIS & 0.44 & 0.54 & 0.45 & 0.55 & 0.38 & 0.56 & - & - & 0.44 & 0.54 \\ 
\hline 
FB15K-237 & 0.31 & 0.56 & 0.29 & 0.56 & 0.17 & 0.42 & 0.22 & 0.51 & 0.35 & 0.59 \\ 
\hline 
YAGO3-10 & 0.19 & 0.32 & 0.15 & 0.31 & 0.22 & 0.27 & 0.16 & 0.26 & 0.20 & 0.34 \\ 
\hline 
DRKG & 0.17 & 0.29 & 0.17 & 0.28 & 0.14 & 0.26 & 0.14 & 0.27 & 0.20 & 0.31 \\ 
\hline 
OPENBIOLINK & 0.28 & 0.32 & 0.24 & 0.27 & 0.47 & 0.52 & 0.27 & 0.30 & 0.28 & 0.32 \\ 
\hline 
\multicolumn{11}{|c|}{top-$k$ = 500} \\ 
\hline 
WN18RR & 0.25 & 0.49 & 0.22 & 0.48 & 0.39 & 0.64 & 0.17 & 0.37 & 0.27 & 0.53 \\ 
\hline 
CARCINOGENESIS & 0.52 & 0.74 & 0.48 & 0.71 & 0.54 & 0.72 & 0.53 & 0.76 & 0.51 & 0.73 \\ 
\hline 
MUTAGENESIS & 0.44 & 0.54 & 0.52 & 0.58 & - & - & - & - & 0.44 & 0.54 \\ 
\hline 
FB15K-237 & 0.31 & 0.56 & 0.29 & 0.56 & 0.17 & 0.42 & 0.22 & 0.51 & 0.35 & 0.59 \\ 
\hline 
YAGO3-10 & 0.19 & 0.32 & 0.15 & 0.31 & 0.22 & 0.27 & 0.16 & 0.26 & 0.20 & 0.34 \\ 
\hline 
DRKG & 0.17 & 0.29 & 0.17 & 0.28 & 0.14 & 0.26 & 0.14 & 0.27 & 0.20 & 0.31 \\ 
\hline 
OPENBIOLINK & 0.28 & 0.32 & 0.24 & 0.27 & 0.47 & 0.52 & 0.27 & 0.30 & 0.28 & 0.32 \\ 
\hline 
\multicolumn{11}{|c|}{top-$k$ = 5000} \\ 
\hline 
WN18RR & 0.25 & 0.49 & 0.16 & 0.45 & 0.06 & 0.85 & 0.14 & 0.38 & 0.30 & 0.54 \\ 
\hline 
CARCINOGENESIS & 0.52 & 0.74 & 0.43 & 0.72 & 0.40 & 0.74 & 0.53 & 0.75 & 0.51 & 0.73 \\ 
\hline 
MUTAGENESIS & 0.44 & 0.54 & 0.44 & 0.49 & 0.27 & 0.31 & - & - & 0.44 & 0.54 \\ 
\hline 
FB15K-237 & 0.31 & 0.56 & 0.29 & 0.56 & 0.19 & 0.44 & 0.22 & 0.50 & 0.35 & 0.59 \\ 
\hline 
YAGO3-10 & 0.19 & 0.32 & 0.15 & 0.32 & 0.16 & 0.40 & 0.16 & 0.26 & 0.20 & 0.34 \\ 
\hline 
DRKG & 0.17 & 0.29 & 0.17 & 0.29 & 0.15 & 0.27 & 0.14 & 0.27 & 0.20 & 0.31 \\ 
\hline 
OPENBIOLINK & 0.28 & 0.32 & 0.25 & 0.29 & 0.45 & 0.54 & 0.24 & 0.27 & 0.29 & 0.33 \\ 
\hline 
\end{tabular}}
\end{table*}

\begin{table*}[h!]
\caption{Average confidence scores using the RotatE model}
\label{tab:rotate-confidence}
\resizebox{\textwidth}{!}{\begin{tabular}{|l|cc|cc|cc|cc|cc|}
 \hline \multicolumn{11}{|c|}{top-$k$ = 50} \\ \hline 
 \multirow{2}{*}{Dataset} & \multicolumn{2}{c|}{\textbf{Rules before}} & \multicolumn{2}{c|}{\textbf{Rules after}} & \multicolumn{2}{c|}{\textbf{New rules}}    & \multicolumn{2}{c|}{\textbf{Dropped rules}}& \multicolumn{2}{c|}{\textbf{Same rules}} \\ \cline{2-11} & \multicolumn{1}{p{1.5cm}}{\centering Std.\ conf.}    & \multicolumn{1}{p{1.5cm}|} {\centering PCA \ conf. }  & \multicolumn{1}{p{1.5cm}}{\centering Std. \ conf.}& \multicolumn{1}{p{1.5cm}|}{\centering PCA \ conf.}  & \multicolumn{1}{p{1.5cm}}{\centering Std.\  conf.} & \multicolumn{1}{p{1.5cm}|} {\centering PCA \  conf.}  & \multicolumn{1}{p{1.5cm}}{\centering Std. \  conf.} &\multicolumn{1}{p{1.5cm}|}{\centering PCA \ conf.} & \multicolumn{1}{p{1.5cm}}{\centering Std. \  conf.} & \multicolumn{1}{p{1.5cm}|}{\centering PCA \ conf.} \\ 
\hline
WN18RR & 0.25 & 0.49 & 0.27 & 0.51 & 0.95 & 1.00 & 0.19 & 0.40 & 0.26 & 0.51 \\ 
\hline 
CARCINOGENESIS & 0.52 & 0.74 & 0.49 & 0.71 & 0.54 & 0.73 & 0.53 & 0.75 & 0.51 & 0.73 \\ 
\hline 
MUTAGENESIS & 0.44 & 0.54 & 0.47 & 0.54 & - & - & - & - & 0.44 & 0.54 \\ 
\hline 
FB15K-237 & 0.31 & 0.56 & 0.29 & 0.56 & 0.21 & 0.42 & 0.22 & 0.51 & 0.35 & 0.59 \\ 
\hline 
YAGO3-10 & 0.19 & 0.32 & 0.15 & 0.31 & 0.22 & 0.27 & 0.16 & 0.26 & 0.20 & 0.34 \\ 
\hline 
DRKG & 0.17 & 0.29 & 0.17 & 0.28 & 0.14 & 0.26 & 0.14 & 0.27 & 0.20 & 0.31 \\ 
\hline 
OPENBIOLINK & 0.28 & 0.32 & 0.24 & 0.27 & 0.42 & 0.47 & 0.27 & 0.30 & 0.28 & 0.32 \\ 
\hline 
\multicolumn{11}{|c|}{top-$k$ = 500} \\ 
\hline 
WN18RR & 0.25 & 0.49 & 0.27 & 0.51 & 0.95 & 1.00 & 0.19 & 0.40 & 0.26 & 0.51 \\ 
\hline 
CARCINOGENESIS & 0.52 & 0.74 & 0.44 & 0.72 & 0.37 & 0.74 & 0.53 & 0.75 & 0.51 & 0.74 \\ 
\hline 
MUTAGENESIS & 0.44 & 0.54 & 0.46 & 0.53 & - & - & - & - & 0.44 & 0.54 \\ 
\hline 
FB15K-237 & 0.31 & 0.56 & 0.30 & 0.56 & 0.28 & 0.45 & 0.22 & 0.51 & 0.35 & 0.59 \\ 
\hline 
YAGO3-10 & 0.19 & 0.32 & 0.15 & 0.31 & 0.16 & 0.26 & 0.15 & 0.26 & 0.21 & 0.35 \\ 
\hline 
DRKG & 0.17 & 0.29 & 0.17 & 0.28 & 0.14 & 0.26 & 0.14 & 0.27 & 0.20 & 0.31 \\ 
\hline 
OPENBIOLINK & 0.28 & 0.32 & 0.24 & 0.27 & 0.40 & 0.45 & 0.25 & 0.28 & 0.29 & 0.33 \\ 
\hline 
\multicolumn{11}{|c|}{top-$k$ = 5000} \\ 
\hline 
WN18RR & 0.25 & 0.49 & 0.46 & 0.53 & 0.48 & 0.54 & 0.18 & 0.44 & 0.28 & 0.51 \\ 
\hline 
CARCINOGENESIS & 0.52 & 0.74 & 0.45 & 0.69 & 0.49 & 0.66 & 0.53 & 0.75 & 0.51 & 0.73 \\ 
\hline 
MUTAGENESIS & 0.44 & 0.54 & 0.39 & 0.45 & - & - & - & - & 0.44 & 0.54 \\ 
\hline 
FB15K-237 & 0.31 & 0.56 & 0.30 & 0.56 & 0.27 & 0.50 & 0.22 & 0.50 & 0.35 & 0.59 \\ 
\hline 
YAGO3-10 & 0.19 & 0.32 & 0.16 & 0.33 & 0.22 & 0.42 & 0.16 & 0.26 & 0.20 & 0.35 \\ 
\hline 
DRKG & 0.17 & 0.29 & 0.16 & 0.28 & 0.13 & 0.27 & 0.14 & 0.27 & 0.20 & 0.31 \\ 
\hline 
OPENBIOLINK & 0.28 & 0.32 & 0.25 & 0.29 & 0.46 & 0.52 & 0.23 & 0.25 & 0.30 & 0.34 \\ 
\hline 
\end{tabular}}
\end{table*}

\subsection{Discussion}
In our preliminary experiments, we compared our proposed approach to AMIE+ and AnyBURL in terms of numbers of rules generated, total runtimes, and performance on link prediction on the test sets.
The results we obtained can be summarized into the following: 1) AnyBURL generates approximately 500$\times$ more rules than AMIE+ and our approach, 2) our proposed approach is slightly slower than AMIE+, but overall faster than AnyBURL, and 3) AnyBURL outperforms our approach on link prediction but only on small datasets as we observed the opposite on large datasets. This suggests that although our approach is based on a simple idea, it can still compete with complex systems in terms of quality of the generated rules.

In our second set of experiments, we investigated the main differences between our approach and AMIE+. Tables~\ref{tab:transe-generated}, \ref{tab:distmult-generated}, and \ref{tab:rotate-generated} give the statistics of different categories of rules, i.e., \emph{before enrichment (AMIE+ rules), after enrichment (rules by our system), new, dropped}, and \emph{same}, using TransE, DistMult, and RotatE embedding-based knowledge completion, respectively. The results suggest that we are able to discover new rules on all knowledge graphs except for TransE and RotatE on MUTAGENESIS. This exception is due to the fact that new triples inferred by the respective embedding models are mostly similar to those already present in the graph, and are hence insufficient to induce new rules.

The average confidence scores of each category are reported in Table~\ref{tab:transe-confidence}, Table~\ref{tab:distmult-confidence}, and Table~\ref{tab:rotate-confidence}, respectively. In each of these tables, we can observe that the average confidence scores remain relatively the same after knowledge graph enrichment. Meanwhile, dropped rules are mostly of low confidence scores except on CARCINOGENESIS. This suggests that our approach can not only discover new rules but also discard low quality rules that become inconsistent with the enriched graph. In the next section, we show treatment-related rules discovered by our approach on DRKG.

\section{Analyzing newly generated rules on DRKG}
\label{sec:analyze-drkg}
In this section, we present and discuss some of the rules generated by our approach on the drug repurposing knowledge graph (DRKG). Without loss of generality, we present rules generated using RotatE embeddings.

Up to 1696 new rules were discovered on this knowledge graph using the RotatE embedding model (Table~\ref{tab:rotate-generated}). Table~\ref{tab:rules} presents 6 rules related to disease treatment in decreasing order of their confidence scores. In the Table, \texttt{GNBR::C::\-Compound:Disease} stands for \textit{inhibits cell growth}, \texttt{GNBR::Pr::Compound:Disease}\\ for \textit{prevents or suppresses}, \texttt{GNBR::Pa::Compound:Di-\\sease} for \textit{alleviates or reduces}, \texttt{GNBR::T::Compound-\\:Disease} for \textit{treats}, \texttt{Hetionet::CrC::Compound:-\\Compound} for \textit{resembles}, and \texttt{INTACT::PHOS\_REACT-\\ION::Gene:Gene} for \textit{phosphorylation reaction}. Rules $R_1$, $R_2$, $R_3$, $R_4$ and $R_6$ relate diseases and chemical compounds. In particular, they suggest sufficient conditions for chemical compounds to be potential candidates to reduce or treat the given diseases. In natural language, $R_6$ can be interpreted as follows: \emph{If a compound is similar to another compound, then they may be able to treat the same disease}. Of course, where one draws the line on the similarity is important, but also not sufficient\textemdash there are other properties to consider for compatibility. $R_5$ relates chemical compounds to genes and can be interpreted as: \emph{If a compound $Y$ inhibits a gene $X$ that has a phosphorylation reaction with a gene $Z$, then $Y$ is an antibody for $Z$}. The two interpretations above clearly show the necessity to involve human experts for further investigations. In this sense, our approach can be used in an initial step to suggest directions for research in specific fields, e.g., drug discovery. 
\begin{table*}[ttb]
\caption{Treatment-related rules on DRKG}
\label{tab:rules}
\resizebox{\textwidth}{!}{\begin{tabular}{|cc|}
\hline
\multicolumn{1}{|c}{Rules} & \multicolumn{1}{c|}{PCA conf.} \\ \hline
 \texttt{DRUGBANK::treats::Compound:Disease}$(X,Y)$ $\land$ \texttt{GNBR::C::Compound:Disease}$(X,Y)$ &\\ ($R_1$)\quad\big\Downarrow & 0.70\\ \texttt{GNBR::Pr::Compound:Disease}$(X,Y)$ &\\ \hline
 
 \texttt{GNBR::Pr::Compound:Disease}$(X, Y)$ $\land$ \texttt{GNBR::T::Compound:Disease}$(X, Y)$ &\\ ($R_2$)\quad\big\Downarrow & 0.69\\ \texttt{DRUGBANK::treats::Compound:Disease}$(X, Y)$ &\\ \hline
 
 \texttt{DRUGBANK::treats::Compound:Disease}$(X, Y)$ $\land$ \texttt{GNBR::C::Compound:Disease} &\\ ($R_3$)\quad\big\Downarrow & 0.67 \\ \texttt{GNBR::Pa::Compound:Disease}$(X, Y)$ &\\ \hline

 \texttt{GNBR::Pa::Compound:Disease}$(X, Y)$ $\land$ \texttt{GNBR::T::Compound:Disease}$(X, Y)$ &\\ ($R_4$)\quad\big\Downarrow & 0.58 \\ \texttt{DRUGBANK::treats::Compound:Disease}$(X, Y)$ &\\ \hline

 \texttt{DGIDB::INHIBITOR::Gene:Compound}$(X, Y)$ $\land$ \texttt{INTACT::PHOS\_REACTION::Gene:Gene}$(X, Z)$ &\\ ($R_5$)\quad\big\Downarrow & 0.40\\ \texttt{DGIDB::ANTIBODY::Gene:Compound}$(Z, Y)$ &\\ \hline

\texttt{DRUGBANK::treats::Compound:Disease}$(X, Y)$ $\land$ \texttt{Hetionet::CrC::Compound:Compound}$(Z, X)$ &\\ ($R_6$)\quad\big\Downarrow & 0.11\\ \texttt{DRUGBANK::treats::Compound:Disease}$(Z, Y)$ &\\ \hline
\end{tabular}}
\end{table*}

\section{Conclusions}
 We investigated whether and how embedding-based knowledge completion on knowledge graphs can help discover new valuable logical rules. To this end, we employed three  different embedding models, TransE, DistMult, and RotatE, to predict potential missing triples in a given knowledge graph. The predicted links between entities are used to enrich the knowledge graph before applying the rule mining system AMIE+. Our experimental results suggest that our approach discovers new rules on seven different datasets. Moreover, rules mined by our approach achieve better link prediction performance on the largest datasets DRKG and OPENBIOLINK. In short, our proposed approach has three main advantages: First, as opposed  to recent expressive rule mining systems such as AnyBURL, our approach is scalable to large datasets as it is based on the fast rule mining system AMIE+ and embedding models. Second, it is able to discover new rules that the original algorithm AMIE+ could not find. Finally, it outperforms expressive rule mining systems on large datasets. Our approach should therefore serve as a strong alternative in situations where expressive rule mining systems are either poor-performing or prohibitively slow.

\section*{Acknowledgments}
This work has received funding from the European Union's Horizon 2020 research and innovation programme under the Marie Skłodowska-Curie grant No 860801 and the European Union’s Horizon Europe research and innovation programme under the grant No 101070305.

\balance
\bibliographystyle{plainnat}
\bibliography{main}
\end{document}

%% file: notation.tex
\newcommand{\kg}{\ensuremath{\mathcal{G}}\xspace}

\newcommand{\ent}{\ensuremath{\mathcal{E}\xspace}}
\newcommand{\rel}{\ensuremath{\mathcal{R}\xspace}}
\newcommand{\tgrel}{\ensuremath{\mathcal{T}\xspace}}